\documentclass[11pt]{article}

\usepackage[preprint]{acl}

\usepackage{times}
\usepackage{latexsym}

\usepackage{amsmath}
\usepackage{amssymb}
\usepackage[T1]{fontenc}

\usepackage[utf8]{inputenc}

\usepackage{microtype}

\usepackage{inconsolata}

\usepackage{graphicx}

\usepackage{hyperref}
\usepackage{amsmath}
\usepackage{url}
\usepackage{array}
\usepackage{todonotes}
\usepackage{booktabs}
\usepackage{adjustbox}
\usepackage{multirow}
\usepackage{makecell}
\usepackage{bbm}

\usepackage{algorithm}
\usepackage{algorithmic}

\title{Tailoring the Curriculum: Student-Centered Reasoning Distillation via Dynamic Data-Model Compatibility}

\author{
  Jiahao Huang$^{1}$ \quad Fei Cheng$^{2,3}$ \quad Junfeng Jiang$^{3}$ \quad Akiko Aizawa$^{1,3}$ \\[2pt]
  $^{1}$University of Tokyo \quad $^{2}$Kyoto University \quad $^{3}$National Institute of Informatics \\[2pt]
  \texttt{jiahao-huang@g.ecc.u-tokyo.ac.jp} \quad \texttt{feicheng@i.kyoto-u.ac.jp} \\
  \texttt{\{jiang, aizawa\}@nii.ac.jp}
}

\begin{document}
\maketitle
\begin{abstract}
Reasoning distillation transfers complex reasoning abilities from large language models (LLMs) to smaller ones, yet its success depends on how well the training data align with the student model. This paper introduces the Data–Model Compatibility (DMC) metric, which can be used to assess the suitability of a dataset for reasoning distillation on a student model. DMC provides an assessment by jointly considering data quality, relative difficulty, and student capability.
We validated the effectiveness of DMC from two perspectives: (1) DMC exhibits a strong correlation with reasoning distillation performance; and (2) using DMC as the criterion for data selection leads to improved reasoning distillation performance. Both findings are consistently demonstrated across multiple student models and tasks.
Moreover, since the DMC of each dataset dynamically changes during training, our experiments demonstrate that dynamically selecting datasets based on DMC can further enhance performance. 
\end{abstract}

\section{Introduction}
\label{sec: intro}

In recent years, a significant number of reasoning models, including OpenAI o1 \citep{jaech2024openai}, DeepSeek-R1 \citep{guo2025deepseek}, and QwQ \citep{qwq32b}, have emerged. These large models have demonstrated outstanding performance on reasoning-dependent tasks such as logic and mathematics. However, the reasoning capabilities of small and medium-sized models remain underdeveloped. Due to their lower resource consumption and higher flexibility, smaller models are more widely adopted in scenarios where both efficiency and effectiveness are required. Therefore, researchers aim to compress the reasoning ability of large models into small ones, which we refer to in this paper as \textbf{reasoning distillation}.

In reasoning distillation, the student models are finetuned on datasets comprising questions, answers, and corresponding reasoning processes generated by the teacher models. 
Prior work \cite{zhang2025quest, xu2025stronger, you2017learning, li2025crowdselect}  primarily focused on how to choose the combination of teacher models and reasoning process generation methods to improve the performance of reasoning distillation. However, we argue that teacher models and generation methods are just indirect factors affecting reasoning distillation, while the most direct factors are the features of the reasoning dataset and the student model. In this paper, we aim to investigate reasoning distillation from a new perspective, focusing on the selection, evaluation, and combination of the features of the dataset and the student model.

\textbf{RQ1}: \textit{Which features of the datasets and student models can effectively reflect the performance of reasoning distillation?}

We analyze features of the dataset and the student model from three perspectives: \textbf{Data Quality $Q$} (a feature of the data), \textbf{Relative Difficulty $D$} (a joint feature of the data and the student model), and \textbf{Student Capability $C$} (a feature of the student model). Their precise definitions and computation are given in Section \ref{sec: foundational features}.

Based on these features, we propose data-model compatibility ($\text{DMC}$), formulated as a function of $Q$, $D$, and $C$, to evaluate the suitability of a dataset for reasoning distillation on a student model. We demonstrate its effectiveness in two ways: (i) DMC values correlate strongly with reasoning-distillation performance across datasets and students; and (ii) constructing datasets from high-DMC data yields better-performing students.



The relative difficulty $D$ and student capability $C$ are naturally dynamic, as they depend on the evolving capacity of the model itself during training. Therefore, the data exhibiting high DMC values will also change dynamically over training. We thus pose the second research question:


\textbf{RQ2}: \textit{Can dynamic data selection according to the evolving DMC values further enhance the performance of reasoning distillation?}


We address two research gaps through this research question. 
First,  we propose an innovative data selection approach for reasoning distillation, adaptively selects the most compatible training data, making the data selection process responsive and tailored to the model’s reasoning level.
Second, compared with traditional dynamic data‑selection methods that rely solely on perplexity \cite{li2024superfiltering, zhang-etal-2025-learning-like}, 
DMC is empirically derived from extensive model–data experiments, which offers a more substantial data‑driven basis.


In summary, the main contributions of this paper are as follows:
(1) We propose data-model compatibility (DMC), a metric jointly modeling three features: data quality, relative difficulty, and student capability, for effectively assessing whether a dataset is suitable for performing reasoning distillation on a student model.
(2) We propose a dynamic data selection approach based on DMC that adaptively re-selects training data to match the student's evolving capability throughout training, effectively enhancing reasoning distillation performance on the test set.

\section{Related Work}
\paragraph{Reasoning Ability of LLMs}

Studies have demonstrated that incorporating a reasoning process into LLMs in question-answering (QA) tasks can enhance model performance \citep{wei2022chain, kojima2022large}. Multiple methods have been proposed to generate reasoning processes in large language models (LLM), such as vanilla chain-of-thought (CoT) \citep{kojima2022large, hsieh2023distilling, mukherjee2023orca, mitra2023orca, lewkowycz2022solving}, tree-of-thought \citep{yao2023tree} , reverse thinking \citep{chen2024reverse}, and self-reflection \citep{li2025dancing, li2025reflectevo}. 


\paragraph{Data Selection}
Recently, with the growing number of data generation methods, increasing attention has been devoted to data selection to further enhance the effectiveness of reasoning distillation. From the perspective of method, \cite{zhang2025quest}, \cite{tian2025not} and \cite{chen2025unveiling} identified teacher model and generation approach as significant factors in reasoning distillation. 
From the perspective of data, quality and difficulty are usually employed. Quality is usually evaluated by LLM evaluators \citep{chen2024genqa, liu2024selectit, lee2024mentor} or reward models \citep{xu2024magpie}. Difficulty can be quantified by approaches such as perplexity (PPL), conditional PPL \citep{li-etal-2024-quantity}, IFD \cite{li2024superfiltering}.
Integration of both metrics like Compatibility-Adusted Reward (CAR)  \citep{xu2025stronger} is also recently introduced.
Unlike these methods, which rely on fixed, hand-designed assumptions (e.g., higher quality or difficulty is always better) and overlook the student model itself, our DMC is data-driven and explicitly conditions on the student's evolving capability, selecting data that best matches the student's current state throughout training.

\section{Preliminaries}
\label{sec: prelim}

\begin{figure*}
    \centering
    \includegraphics[width=0.8\linewidth]{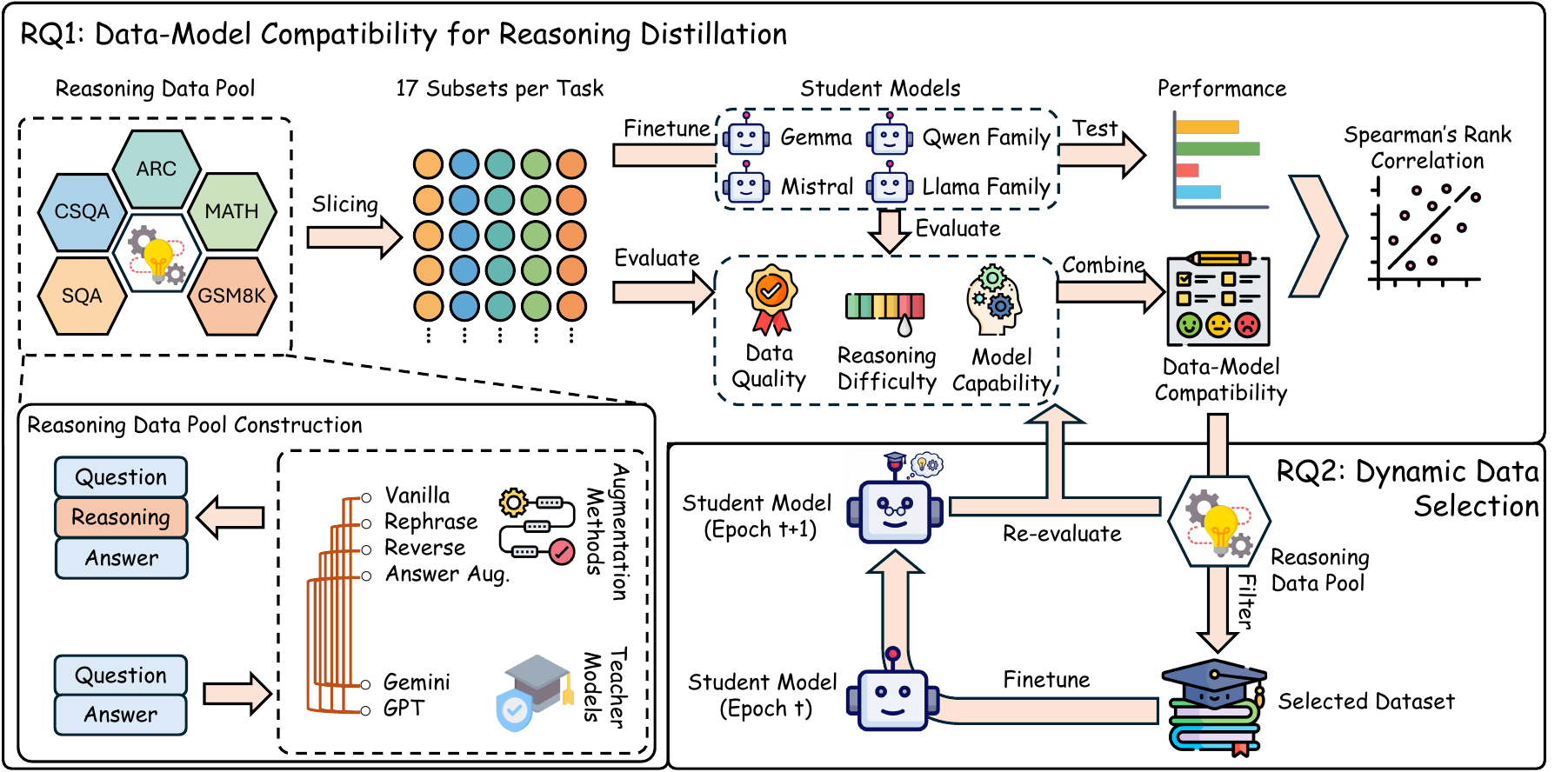}
    \caption{Pipeline of this paper. The DMC metric is formulated in Section \ref{sec: method}; both research questions are empirically validated in Section \ref{sec: results}.}
    \label{fig:pipeline}
\end{figure*}

\subsection{Problem Formulation}
\label{sec: problem formulation}
Fig. \ref{fig:pipeline} illustrates the formulation of this research problem as well as some of its specific settings. Reasoning distillation starts from a raw question-answering (QA) dataset $\mathcal{D}_0=\{(q,a)\}$, where $q$ is the question and $a$ is the ground truth answer. 
For each question-answer pair, the teacher model $T$ employs an augmentation method $Aug$ to generate a reasoning process $r$. Consequently, the entire dataset can be expanded into a reasoning dataset $\mathcal{D}(T,Aug)=\{(q,r,a)\}$. To ensure diversity in the reasoning processes, multiple teacher models and augmentation methods can be employed. We aggregate all the reasoning datasets into a reasoning data pool: $U = \bigcup_{T,Aug}\mathcal{D}(T,Aug)$.

For comparative purposes, we sample several subsets $\mathcal{D}_i \subseteq U$ from the data pool. 
On one hand, we finetune the student model $S$ on $\mathcal{D}_i$ and denote its test set performance by $P_{S}(\mathcal{D}_i)$. 
On the other hand, we aim to identify a metric $M$ to evaluate the suitability of employing $\mathcal{D}_i$ for reasoning distillation on the student model $S$, denoted as $M_{S}(\mathcal{D}_i)$. High correlation between $[M_{S}(\mathcal{D}_i)]|_i$ and $[P_{S}(\mathcal{D}_i)]|_i$ indicates that $M$ serves as an effective metric.

\subsection{Foundational Features}
\label{sec: foundational features}

Whether a dataset benefits reasoning distillation is not an intrinsic property of the data alone; it depends on how well the data fits the particular student model. We therefore characterize this fit along three complementary features. \textbf{Data quality} ($Q$) captures the data side: a reasoning chain that is incorrect, incoherent, or fails to reach the answer teaches the student flawed patterns no matter which model is trained on it. \textbf{Relative difficulty} ($D$) captures the interaction between the data and the model: a chain that is too hard cannot be absorbed by the current student, whereas one that is too easy conveys little new signal, so what matters is the difficulty \emph{relative} to the student. \textbf{Student capability} ($C$) captures the model side: since the same data suits different students to different degrees, a metric that ignores the student cannot express compatibility at all. Together, $Q$, $D$, and $C$ cover the data, the model, and their interaction, which makes them a natural basis for analyzing data-model compatibility (DMC). In this subsection we describe how each feature is computed; how they are integrated into the DMC metric is presented in Section \ref{sec: method}.

\paragraph{Data Quality $Q$}
Data quality is a metric for evaluating whether the reasoning process $r$ is correct, accurate, coherent and leading to the ground truth answer. Following \citet{xu2025stronger, xu2024magpie}, we employ state-of-the-art reward models to score the quality of each single data entry; the specific reward models are detailed in Section \ref{sec: experiment setup}. For each subset $\mathcal{D}_i$, we define the quality of the dataset as the average quality of all data entries within it. Since data quality depends only on the data, the subscript for student model $S$ is omitted and denoted as $Q(\mathcal{D}_i)$.


\paragraph{Relative Difficulty $D$}
We emphasize that difficulty here is not an intrinsic property of the data but is always defined \emph{relative to a given student model}: it measures to what extent that particular model finds the reasoning chain hard to comprehend, so the same chain can be easy for a strong student and hard for a weak one. Perplexity (PPL), conditional perplexity (CPPL) and instruction following difficulty (IFD) \cite{li2024superfiltering} are the mainstream metrics for evaluating this relative difficulty. For student model $S$, PPL of a reasoning $r$ can be calculated by
\begin{align}
\text{PPL}_S(r)=\exp(-\frac{1}{N}\sum_{i=1}^N\log p_S(r^i|r^{1:i-1}))
\end{align}
where $r^i$ represents the $i$-th token of $r$ and $N$ is the number of tokens in $r$. CPPL can be calculated by $\text{PPL}_S(r|q)$
and IFD is defined as the ratio of CPPL and PPL. 
Similar to $Q$, the difficulty of dataset $\mathcal{D}_i$ for student model $S$ is defined as the average difficulty of all data samples in $\mathcal{D}_i$.


\paragraph{Student Capability $C$}
Previous work mainly focused on the features of data while disregarding the model itself; therefore, we introduce student capability $C$ as another key factor in the formulation of DMC.

The concept of student capability is inspired by the placement test in human education, which first evaluates a student’s capability and then designs a suitable training course.
From the data pool $U$, we isolate a small subset $\mathcal{D}_p$ of high-quality samples to serve as a placement test for the preliminary evaluation of the student capability; these samples are held out from subsequent training, and the sampling details are given in Section \ref{sec: experiment setup}. We consider student capability as the extent to which the model can comprehend the reasoning data from the placement test. Therefore, we define the absolute capability value of the student model $S$ as:
\begin{align}
C_S^{abs}=\mathbb{E}_{(q,r)\in D_p}\frac{1}{N}\sum_{i=1}^N\log p_S(r^i|q, r^{1:i-1})
\end{align}
A higher value of $C_S^{abs}$ indicates greater confidence and familiarity with high-quality reasoning in the placement test, thereby reflecting stronger capability.
By linearly mapping absolute capability values of different student models on various raw datasets $\mathcal{D}_0$ to the range ([0,1]), we can obtain corresponding relative capability value $C_S^{rel}$.

\section{Method: Data-Model Compatibility}
\label{sec: method}

\subsection{DMC Formulation}
\label{sec: dmc formulation}
Instead of directly measuring the features of the data, student capability $C$ plays the role of a \emph{modulator}: we adopt two evaluation metrics, $M_S^L(\cdot)$ and $M_S^H(\cdot)$, designed for student models with the lowest and highest capability respectively, and let the modulator smoothly transition between them. Providing different evaluation criteria to students of different capability resembles how a teacher tailors a curriculum to each student's level: what matters is not the sheer difficulty of the data, but its compatibility with the student.

Given a data entry $d$ and a student model $S$, to allow DMC to adapt its evaluation metric based on the capability of the model, we employ a nonlinear interpolation method to model DMC:
\begin{multline}
\label{eqa: model dmc}
    \text{DMC}_S(d;Q,D,C,M_S^L(\cdot),M_S^H(\cdot),f(\cdot)) =
    \\(1-f(C_S))*M_S^L(Q(d),D_S(d))
    \\+f(C_S)*M_S^H(Q(d),D_S(d))
\end{multline}
Here, $M_S^L(\cdot)$ and $M_S^H(\cdot)$ are the two limiting evaluation metrics that DMC interpolates between, corresponding respectively to the lowest- and highest-capability regimes of the student. Both are functions of the quality $Q(d)$ and difficulty $D_S(d)$ of the data entry $d$ with respect to student $S$. The interpolation function $f(C_S) \in [0,1]$ controls the relative influence of each term, enabling a smooth, nonlinear transition from $M_S^L$ to $M_S^H$ as the student’s capability $C_S$ increases. The overall DMC for a dataset $\mathcal{D}_i$ with respect to model (S) is computed as the average DMC across all entries $d \in \mathcal{D}_i$.

\subsection{Optimization Procedure}
\label{sec: optimization}
Our objective is to determine the optimal configuration for Equation \ref{eqa: model dmc}, including the selection of evaluation metric for $Q$, $D$, and $C$, the functional forms of $M_S^L(\cdot)$ and $M_S^H(\cdot)$, and the interpolation function $f(\cdot)$. To ensure generalizability, we aim to maximize the average value of this correlation across different student models and initial datasets:

\begin{multline}
\label{equ: def dmc}
\text{DMC}^{\text{opt}} = \text{argmax}_{Q(\cdot),D(\cdot),C_S,M_S^L(\cdot),M_S^H(\cdot),f(\cdot)}\\
\mathbb{E}_{S,\mathcal{D}_0} Corr([\text{DMC}_{S}(\mathcal{D}_i)]|_i, [P_{S}(\mathcal{D}_i)]|_i)
\end{multline}

To obtain the optimal configuration for $\text{DMC}$, we enumerate all combinations of candidate evaluation metrics for $Q$, $D$, and $C$, and derive the functional forms of $M_S^L(\cdot)$, $M_S^H(\cdot)$ and $f(C_S)$ via symbolic regression. Concretely, the procedure consists of four steps: (1) enumerate all combinations of candidate metrics for $Q$, $D$, and $C$; (2) for each combination, run symbolic regression to obtain candidate functional forms on the Pareto frontier balancing correlation against complexity; (3) filter the candidates to retain those that are human-interpretable; and (4) apply grid search over the remaining constants and select the configuration with the highest correlation. The detailed pseudocode is given in Algorithm \ref{alg:dmc_optimization}, Appendix \ref{app: psuedocode}. The discovered configuration and its empirical analysis are reported in Section \ref{sec: discovered config}.

\subsection{DMC-Based Data Selection}
\label{sec: dmc selection}
Once DMC is obtained, it can be used as a criterion for selecting training data. We consider two strategies, both evaluated in Section \ref{sec: rq2 results}.

\paragraph{Static Selection} Datasets are fixed prior to training and remain unchanged throughout. For each student model $S$, we select the top $k\%$ of data samples from the data pool $U$ according to $\text{DMC}_S$, and use them for reasoning distillation.

\paragraph{Dynamic Selection} Datasets are adjusted throughout training according to the evolving model parameters. Before each training epoch, all data in $U$ are re-evaluated with $\text{DMC}_S$, and the top $k\%$ of samples are selected as the training data for that epoch. Since data quality $Q$ is static, the dynamic behavior is driven by the relative difficulty $D_S$ and student capability $C_S$, both of which evolve as the student learns.


\section{Experimental Setup}
\label{sec: experiment setup}
\paragraph{Data Pool $U$} The data pool $U$ is constructed based on DC-CoT \citep{zhang2025quest}. DC-CoT is a data-centric benchmark that collects reasoning process generated with various teacher models and augmentation strategies. The selection of teacher model $T$ includes Gemini-1.5-Pro \citep{team2024gemini} and GPT-4 \citep{achiam2023gpt}. The selection of augmentation strategy $Aug$ is comprised of vanilla CoT, rephrase questions \citep{yu2023metamath}, reverse thinking \citep{chen2024reverse} and answer augmentation \citep{yu2023metamath}. 

We will focus on the augmentation of the following raw datasets $\mathcal{D}_0$ from DC-CoT: \textit{Commonsense Reasoning}: StrategyQA (SQA) \citep{geva2021did}, CommonsenseQA (CSQA) \citep{talmor2018commonsenseqa}, ARC-challenge (ARC) \citep{clark2018think}; \textit{Math Reasoning}: GSM8K \citep{cobbe2021training}, MATH \citep{hendrycks2021measuring}. These datasets differ in question formats: SQA are comprised of yes/no questions, CSQA and ARC are comprised of multiple-choice questions, and GSM8K and MATH are comprised of arithmetic problems.

\paragraph{Quality Evaluation} To instantiate the data quality $Q$ (Section \ref{sec: foundational features}), we follow \citet{xu2025stronger, xu2024magpie} and consider two state-of-the-art reward models as candidates: \textit{Skywork-Reward-V2-Llama-3.1-8B} (SRL) \citep{liu2025skywork} and \textit{Skywork-Reward-Gemma-2-27B} (SRG) \citep{liu2024skywork}.

\paragraph{Method for Sampling $\mathcal{D}_i$}
From each data pool $U$, we sampled 17 representative datasets $\mathcal{D}_{0:16}$. First, the non-CoT datasets $\mathcal{D}_{0}=\{(q,None,a)\}$ are included as baselines. The sampling method for dataset $\mathcal{D}_{1:8}$ follows the previous work, where each dataset corresponds to a combination of a teacher model and a data‑augmentation method. $\mathcal{D}_{9}$ is equivalent to the full data pool $U$, while $\mathcal{D}_{10}$ is a random sample from the full dataset, with a sample size equal to $\mathcal{D}_{0}$. 
Then, we sorted the data in $U$ based on their quality scores and partitioned them into three equally sized datasets: high-quality, medium-quality, and low-quality. We independently sampled two subsets of the same size as $\mathcal{D}_{0}$ from each of the three data pools, thereby constructing $\mathcal{D}_{11:16}$. These sampled datasets cover a range of teacher models, augmentation strategies, as well as diversity and quality levels, which ensures that they are representative. 

\paragraph{Student Model $S$}
For a comprehensive comparison, we will finetune student models from four model families with different parameter sizes: Gemma-2B \citep{team2024Gemma}; Mistral-7B \citep{jiang2023Mistral7b}; \textit{Qwen}: Qwen2.5-1.5B, Qwen2.5-3B, Qwen2.5-7B \citep{qwen2025qwen25technicalreport}; \textit{Llama3}: Llama-3.2-1B, Llama-3.2-3B, Llama-3.1-8B \citep{grattafiori2024llama3herdmodels}.

\paragraph{Placement Test} To estimate the student capability $C$ (Section \ref{sec: foundational features}), the placement set $\mathcal{D}_p$ consists of 100 samples drawn from the top 10\% highest-quality data, which are held out from training. This small, high-quality probe keeps the estimate efficient and stable while preserving the diversity of the training pool.

\paragraph{Finetuning and Test Set Performance $P$}
The finetuning is conducted on four A100 GPUs, with LoRA finetuning for the student models. The LoRA rank is set as 8, while LoRA alpha is set as 32. The total number of training steps for all datasets is set as 30k, with a batch size of 32 and a learning rate of $1*10^{-5}$. The prompt for finetuning is shown in Table \ref{tab: prompt}, Appendix \ref{app: prompt}. After finetuning, we evaluate the finetuned model on the corresponding test set and adopt accuracy as the criterion for evaluating the performance.

\paragraph{Correlation Calculation} We follow previous work \cite{xu2025stronger} and use Spearman's rank correlation coefficient $\rho$ \cite{zar2005spearman} to evaluate the consistency in relative ranking between a given metric and the test-set performance, which can be calculated as: 
\begin{align}
\rho=1-\frac{6\sum_i d_i^2}{n(n^2-1)}
\end{align}
where $d_i$ is the difference between the rank of $M_{S}(\mathcal{D}_i)$ and $P_{S}(\mathcal{D}_i)$. The domain of $\rho$ ranges from -1 to 1, where 1 indicates a perfect positive correlation between two variables. Our target is to find out the $M$ which maximizes $\rho([M_{S}(\mathcal{D}_i)]|_i, [P_{S}(\mathcal{D}_i)]|_i)$ across different raw datasets and student models.

\paragraph{Baselines} 
To evaluate the effectiveness of DMC, we use data quality and relative difficulty as baselines for comparison. Moreover, we employ compatibility-adjusted reward (CAR) \citep{xu2025stronger}, a multi‑metric integration metric, as an additional baseline.

\section{Results and Analysis}
\label{sec: results}

\subsection{Discovered DMC Configuration}
\label{sec: discovered config}
Running the optimization procedure of Section \ref{sec: optimization} over all candidate metrics and functional forms yields the configuration shown in Table \ref{tab:config DMC}. For brevity, subsequent mentions of DMC denote this optimal configuration. The visualizations of $M_S^L(\cdot)$, $M_S^H(\cdot)$ and $f(C_S)$ are provided in Figure \ref{fig:vis msl msh} and \ref{fig:f(C_S)}, Appendix \ref{app: visualization DMC config}.

\begin{table}
\centering
\adjustbox{max width=0.42\textwidth}{
\begin{tabular}{
>{\raggedright\arraybackslash}m{1.5cm}|
>{\centering\arraybackslash}m{1.2cm}|
>{\centering\arraybackslash}m{5cm}}
\toprule
\multirow{3}{*}{\makecell[c]{\textbf{Metric}\\ \textbf{Selection}}}
& $Q$ & SRL \\
& $D$ & $\log({\text{IFD}})$\\
& $C$ & $C_S^{rel}$\\

\midrule
\multirow{3}{*}{\makecell[c]{\textbf{Function}\\ \textbf{Form}}}
& $M_S^L(\cdot)$ & $\mathbb{I}(Q>Q_{th})*\mu_1*D+\mu_2*\sqrt{Q})$\\
& $M_S^H(\cdot)$ & $\mu_3*Q*e^{-\mu_4(D-D_{base})}$\\
& $f(C_S)$ & $C_S^2/(C_S^2+(1-C_S)^2)$ \\
\midrule
\multirow{6}{*}{\makecell[c]{\textbf{Param.}}}
&$Q_{th}$ & 1.1\\
&$D_{base}$ & 0.056\\
&$\mu_1$ & 2.1\\
&$\mu_2$ & 5.0\\
&$\mu_3$ & 5.0\\
&$\mu_4$ & 0.10\\

\bottomrule
\end{tabular}
}
\caption{Optimal configuration for DMC. SRL is an abbreviation for \textit{Skywork-Reward-V2-Llama-3.1-8B}.}
\label{tab:config DMC}
\end{table}

\subsection{Correlation with Distillation Performance}
\label{sec: rq1 results}
\paragraph{Effectiveness of DMC}
Table \ref{tab:corr-avg} presents the average correlation between $[M_{S}(\mathcal{D}_i)]|_i$ and $[P_{S}(\mathcal{D}_i)]|_i$, while the complete data are provided in Table \ref{tab:corr-full}, Appendix \ref{sec: corr}. Our proposed DMC exhibits a high correlation with $P_{S}(\mathcal{D})$ across various student models $S$ and multiple raw datasets $\mathcal{D}_0$, attaining the highest average correlation of $0.612$ and ranking first on every one of the eight student models. This clearly surpasses the strongest single-metric baseline, data quality (SRL, $0.555$), and the multi-metric baseline CAR ($0.465$), demonstrating that DMC is an effective metric for assessing whether a dataset $\mathcal{D}_i$ is suitable for reasoning distillation with a given student model $S$.

Data quality also serves as a strong metric, as many previous work has pointed out. Higher-quality data naturally enable the student model to perform reasoning of higher quality, thereby increasing the likelihood of producing correct answers. In contrast, relative difficulty is not a consistently effective metric. We find that its effectiveness depends on the student capability (see Table \ref{tab:capability-model-task} for reference). For student models with relatively low capability, such as Gemma‑2B and Mistral‑7B, both CPPL and IFD can serve as effective evaluation metrics. However, once the student capability increases, the relationship between performance and difficulty, particularly IFD, ceases to be purely positively correlated, as shown in Figure \ref{fig:relation pqd}, Appendix \ref{sec: relation pqd}. 

\begin{table*}
\centering
\adjustbox{max width=0.8\textwidth}{
\begin{tabular}{>{\raggedright\arraybackslash}m{1.4cm}
                >{\raggedright\arraybackslash}m{0.9cm}|
                >{\centering\arraybackslash}m{1.2cm}
                >{\centering\arraybackslash}m{1.2cm}
                >{\centering\arraybackslash}m{1.2cm}
                >{\centering\arraybackslash}m{1.2cm}
                >{\centering\arraybackslash}m{1.2cm}
                >{\centering\arraybackslash}m{1.2cm}
                >{\centering\arraybackslash}m{1.2cm}
                >{\centering\arraybackslash}m{1.2cm}|
                >{\centering\arraybackslash}m{1.2cm}}
\toprule
\multicolumn{2}{c|}{\multirow{2}{*}{\makecell[c]{\textbf{Metric} $M$} }}& \multicolumn{8}{c|}{\textbf{Student Models} $S$}& \multirow{2}{*}{\makecell[c]{\textbf{Avg.}}} \\
&&G-2b&M-7b&Q-1.5b&Q-3B&Q-7B&L-1b&L-3b&L-8b& \\
\midrule
\multirow{2}{*}{\makecell[c]{\textbf{Quality}}}
&\textbf{SRL}&0.452 & 0.440 & \underline{0.650} & \underline{0.656} & \underline{0.763} & \underline{0.481} & \underline{0.662} & 0.339 & \underline{0.555} \\
&\textbf{SRG}&0.232 & 0.309 & 0.538 & 0.612 & 0.568 & 0.538 & 0.474 & 0.330 & 0.450 \\

\midrule
\multirow{3}{*}{\makecell[c]{\textbf{Difficulty}}}
&\textbf{PPL}&-0.513 & -0.463 & -0.677 & -0.512 & -0.445 & -0.469 & -0.722 & -0.590 & -0.549 \\
&\textbf{CPPL}&\underline{0.499} & 0.433 & 0.397 & 0.202 & 0.135 & 0.275 & 0.439 & \underline{0.385} & 0.346 \\
&\textbf{IFD}&0.475 & \underline{0.444} & 0.098 & -0.050 & -0.024 & 0.050 & -0.014 & -0.116 & 0.108 \\

\midrule
\multirow{2}{*}{\makecell[c]{\makecell[l]{\textbf{Multi-}\\\textbf{Metric}}}}
&\textbf{CAR}&0.266 & 0.408 & 0.511 & 0.513 & 0.582 & 0.411 & 0.651 & 0.380 & 0.465 \\
&\textbf{DMC}&\textbf{0.539} & \textbf{0.539} & \textbf{0.702} & \textbf{0.669} & \textbf{0.777} & \textbf{0.546} & \textbf{0.735} & \textbf{0.411} & \textbf{0.612}\\
\bottomrule
\end{tabular}
}
\caption{Average correlation between $P_S(\mathcal{D}_i)$ and $M_S(\mathcal{D}_i)$ across different raw datasets. The highest correlation for each model is highlighted in bold, and the second-highest is highlighted with \underline{underline}.}
\label{tab:corr-avg}
\end{table*}

\paragraph{Interpretation of DMC Configuration}
Beyond predicting performance, these results also justify \emph{why} we model DMC with the three features and the interpolation form of Equation \ref{eqa: model dmc}. Figure \ref{fig: dmc-capability-trend} (Appendix \ref{app: visualization DMC config}) shows, for representative settings, that as the student capability $C$ increases, both the best-performing subsets $\mathcal{D}_i$ and the high-DMC region shift from high-difficulty toward moderate-difficulty data. This shift supports our design in two ways. (i)~\emph{Three features.} Because the difficulty that benefits a student depends on \emph{which} student, a metric defined on the data alone (quality and difficulty) cannot express this dependence, which makes the student capability $C$ indispensable alongside $Q$ and $D$; meanwhile, the consistently high correlation of quality across models (Table \ref{tab:corr-avg}) confirms that $Q$ should remain a core term. (ii)~\emph{Interpolation form.} Because the preferred difficulty changes continuously with $C$ rather than being fixed, a single static scoring function is inadequate; this is exactly what the capability-modulated interpolation of Equation \ref{eqa: model dmc} captures, blending a low-capability metric $M_S^L$ (favoring high-quality, high-difficulty data) and a high-capability metric $M_S^H$ (favoring moderate-difficulty data) through $f(C_S)$.

Concretely, low-capability models tend to benefit from high-quality and high-difficulty data, while high-capability models are better suited to data of moderate difficulty. This runs opposite to the curriculum-learning intuition that data difficulty should increase with capability \cite{bengio2009curriculum}. We attribute the reversal to the nature of reasoning distillation: since the student already possesses basic language understanding at initialization, low-capability models mainly lack the ability to \emph{generate} coherent reasoning chains, so challenging examples quickly activate this ability; high-capability models, in contrast, have stable reasoning abilities that overly difficult data may destabilize, making medium-difficulty data the better trade-off between challenge and stability.


The strong correlation observed between $[\text{DMC}_{S}(\mathcal{D}_i)]|_i$ and $[P_{S}(\mathcal{D}_i)]|_i$ confirms that DMC effectively predicts the suitability of a dataset for reasoning distillation, thereby answering RQ1. We next use DMC as a selection criterion to verify its effectiveness from the perspective of downstream distillation performance, addressing RQ2.


\subsection{Data Selection Results}
\label{sec: rq2 results}
We now examine to what extent employing DMC as a data-selection criterion improves reasoning distillation, under both the static and dynamic strategies defined in Section \ref{sec: dmc selection}. For the dynamic setting we additionally compare against the dynamic curriculum-learning method ADCL \cite{zhang-etal-2025-learning-like}; all other training settings follow Section \ref{sec: experiment setup}.
In the main text, we use Qwen2.5‑3B as a representative example to illustrate the performance of this student model when conducting reasoning distillation on different datasets, which is listed in Table \ref{tab:performance-q-3b}. The performance of the remaining models are provided in Appendix \ref{app: performance}.

\begin{table}
\centering
\adjustbox{max width=0.5\textwidth}{
\begin{tabular}{
>{\raggedright\arraybackslash}m{1.8cm}|
>{\raggedright\arraybackslash}m{3.0cm}|
>{\centering\arraybackslash}m{1.2cm}
>{\centering\arraybackslash}m{1.2cm}
>{\centering\arraybackslash}m{1.2cm}
>{\centering\arraybackslash}m{1.2cm}
>{\centering\arraybackslash}m{1.2cm}}
\toprule
\multirow{2}{*}{\makecell[c]{\makecell[l]{\textbf{Training}\\\textbf{Dataset}}}} & \multirow{2}{*}{\makecell[c]{\makecell[l]{\textbf{Selection Metric}}\\\textbf{(Top $k\%$)}}} & \multicolumn{5}{c}{\textbf{Tasks}}\\
& & SQA & CSQA & ARC & MATH & GSM8K \\
\midrule
\textbf{None} & \makecell[c]{-} & 53.9 & 41.0 & 63.2 & 10.2 & 16.4 \\
\midrule
\multirow{9}{*}{\makecell[l]{\textbf{Static}}}
& Full (100\%)  & 59.8 & 68.2 & 75.8 & 11.4 & 71.0 \\
& Random (12.5\%) & 57.2 & 70.8 & 72.2 & 10.6 & 64.8 \\
& Quality (12.5\%) & 66.8 & \underline{71.8} & 76.0 & 56.8 & 73.4\\
& Difficulty (12.5\%) & 52.8 & 69.4 & 64.0 & 21.2 & 50.6\\
& CAR (12.5\%)  & 56.8 & 70.6 & 75.8 & 49.4 & 73.8\\
& DMC (12.5\%)  & 67.0 & \textbf{72.1} & \textbf{77.4} & \textbf{66.4} & \underline{75.6}\\
& DMC (10\%)  & \textbf{69.0} & 71.6 & 76.4 & \underline{64.8} & \textbf{77.4}\\
& DMC (7.5\%)  & \underline{67.5} & 71.4 & \underline{77.0} & 58.6 & 73.8\\
\midrule
\multirow{6}{*}{\textbf{Dynamic}}
& ADCL & 62.9 & 69.6 & 65.4 & 37.0 & 57.8\\
& Difficulty (12.5\%) & 48.5 & 68.2 & 65.8 & 34.8 & 50.6\\
& CAR (12.5\%) & 57.2 & 70.6 & 75.8 & 55.6 & 76.6\\
& DMC (12.5\%) & \underline{68.5} & 72.6 & \textbf{79.6} & \textbf{69.0}& \textbf{79.2} \\
& DMC (10\%) &  \textbf{69.1} & \underline{73.4} & \underline{78.2 } & 66.0 & 78.2\\
& DMC (7.5\%) & 68.2 & \textbf{73.8} & 77.4 & \underline{67.4} & \underline{78.8}\\
\bottomrule
\end{tabular}
}
\caption{Performance of Qwen2.5-3B. For the static and dynamic settings, the dataset selection metric achieving the best and second-best performance are indicated in \textbf{bold} and \underline{underline}, respectively.}
\label{tab:performance-q-3b}
\end{table}

As shown, in reasoning distillation, the use of the full dataset does not lead to improved performance, suggesting that the scaling law no longer applies in this task; on MATH, for instance, the full dataset reaches only $11.4$, barely above the $10.2$ of the non-distilled model. The reason is that at this stage, the student model already possesses a certain level of capability in understanding and processing natural language, making the training process more sensitive to data‑model compatibility. Consequently, effective data selection has already become a crucial task.

From the perspective of the metrics, DMC serves as a superior metric for data selection compared with others. This advantage arises because DMC not only takes into account the inherent quality of the data but also considers the difficulty of the data as perceived by the student model and the model’s capability, effectively embodying a “personalized learning” principle.  This improvement is particularly evident in tasks such as GSM8K and MATH, where reasoning ability plays a more critical role: on MATH, dynamic DMC (top $12.5\%$) attains $69.0$, far surpassing the full dataset ($11.4$), CAR ($55.6$), and ADCL ($37.0$). Consistent with Section \ref{sec: rq1 results}, higher difficulty is not always better: especially for high-capability students, stabilizing already-acquired reasoning skills matters more than exploring harder data.

With dynamic data selection, employing DMC as the evaluation criterion yields additional improvement in model performance. Re‑evaluating the data before each epoch allows the dataset to remain aligned with the student model’s current parameter state, thereby continuously enhancing performance. Furthermore, under the same  $k$ value, such re‑evaluation gives the model access to new data, increasing data diversity during training and further improving generalization ability. Other metrics, however, are comparatively simplistic, and therefore show limited benefit even when used with dynamic dataset strategies. The selection ratio $k$ is itself a trade-off: a larger $k$ adapts less promptly to DMC, whereas a smaller $k$ requires more frequent re-evaluation and yields a less stable selection.

In summary, dynamic DMC-based selection effectively enhances reasoning distillation; we next analyze how the model and the selected data evolve during training.

\subsection{Analysis}
\label{sec: discussion}
Under the dynamic setting, student capability, the selected dataset, and relative difficulty all evolve together during training; we use MATH as $\mathcal{D}_0$ with Qwen2.5-3B (high-capacity) and Gemma-2B (low-capacity) as representatives. Two dynamics are intuitive and deferred to Appendix \ref{app: training dynamics}: the student capability rises steadily as training proceeds, and the relative difficulty of individual samples shifts over time so that DMC keeps a sample only while it lies within the effective compatibility range. We highlight here how the selected data is distributed.

\begin{figure}
    \centering
    \includegraphics[width=1\linewidth]{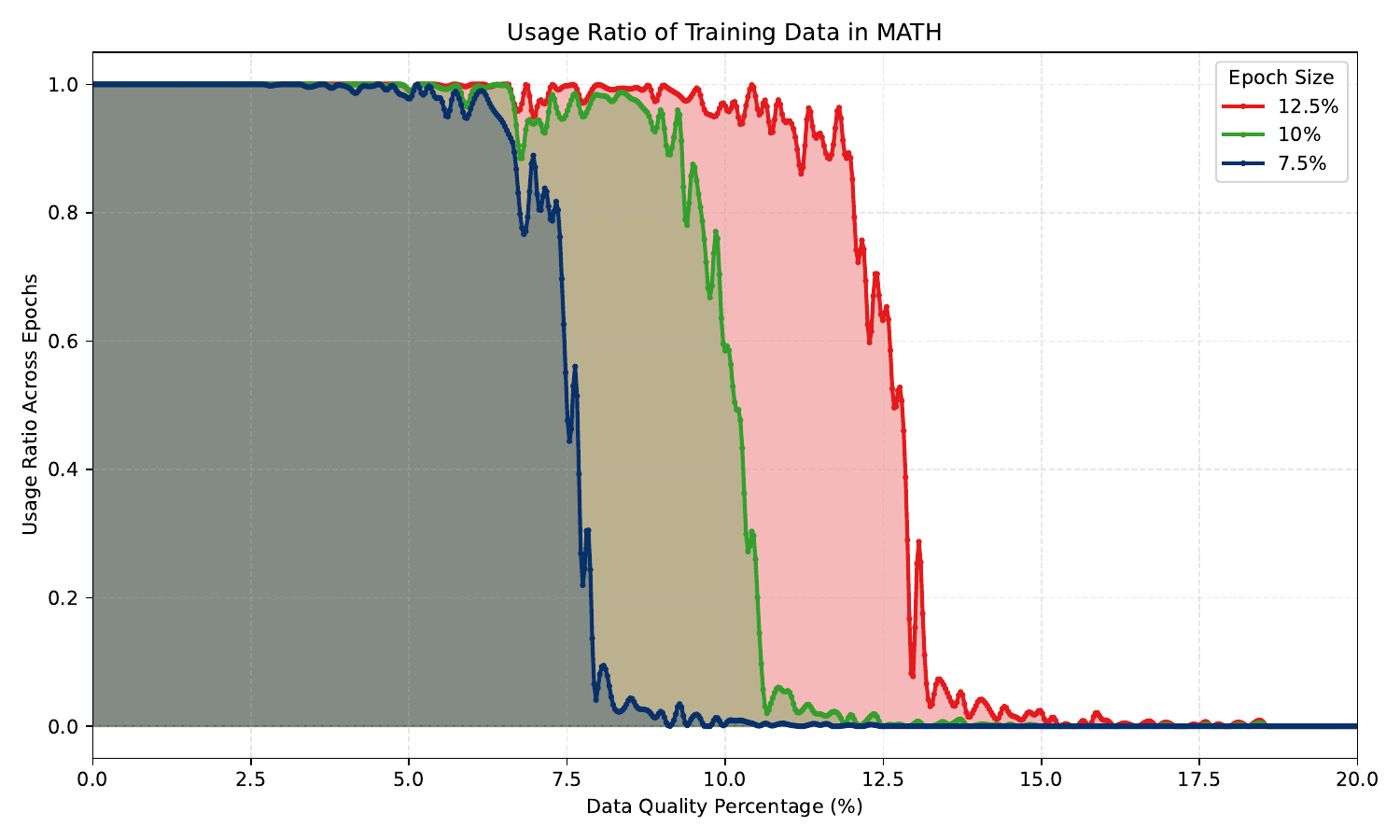}
    \caption{Distribution of data selection frequency, with student model chosen as Gemma-2b, raw dataset $\mathcal{D}_0$ chosen as MATH.}
    \label{fig:distribution selected data}
\end{figure}

\paragraph{Distribution of Selected Data}
We rank all data by descending quality and observe the frequency distribution of data selection over the course of training, as shown in Figure \ref{fig:distribution selected data}. The highest-quality 5\% of data are nearly always included, acting as compulsory ``core courses'' that establish the foundation of the model's primary reasoning ability and must be consistently retained. Data ranked between the top 5\% and 20\% are dynamically adjusted with the model's training progress; this portion acts as ``elective courses'' that flexibly tune reasoning generalization and prevent overfitting on the core data, and its range can be controlled via the epoch size to trade off generalization against reasoning quality. Data beyond this range, being of relatively low quality, are seldom selected. This pattern mirrors human curriculum design and underlies the consistent gains of dynamic DMC selection.

\paragraph{Out-of-Distribution Generalization}
\label{sec: ood}
Finally, we test whether DMC captures a transferable notion of learnability rather than a heuristic specific to math and commonsense reasoning, by applying our selection method to two out-of-distribution (OOD) tasks that differ substantially from the data pool: ANLI \citep{nie2020adversarial} (natural language inference) and Date Understanding \citep{srivastava2023beyond} (temporal reasoning), again with Gemma-2B as the representative student. As shown in Table \ref{tab:ood}, dynamic DMC-based selection (top 12.5\%) outperforms both the full-dataset baseline and the dynamic baselines (ADCL, Difficulty, CAR) on both OOD tasks, indicating that the data-model compatibility signal generalizes beyond the domains used to derive it.

\begin{table}
\centering
\adjustbox{max width=0.38\textwidth}{
\begin{tabular}{
>{\raggedright\arraybackslash}m{3.0cm}|
>{\centering\arraybackslash}m{1.2cm}
>{\centering\arraybackslash}m{1.2cm}}
\toprule
\multirow{2}{*}{\makecell[c]{\makecell[l]{\textbf{Selection Metric}}\\\textbf{(Top $k\%$)}}} & \multicolumn{2}{c}{\textbf{Tasks}}\\
& ANLI & Date \\
\midrule
Full (100\%) & 35.42 & 61.80 \\
ADCL & 43.92 & 60.37 \\
Difficulty (12.5\%) & 42.47 & 67.02 \\
CAR (12.5\%) & 40.29 & 68.41 \\
DMC (12.5\%) & \textbf{49.75} & \textbf{70.41} \\
\bottomrule
\end{tabular}
}
\caption{Out-of-distribution accuracy ($\%$) on ANLI and Date Understanding with Gemma-2B as the student. Selection methods are under the dynamic setting (top 12.5\%) and compared against the full-dataset baseline; the best result per task is in \textbf{bold}.}
\label{tab:ood}
\end{table}

\section{Conclusion}
In this paper, we reveal that effective reasoning distillation depends on a nuanced interplay between data quality, relative difficulty, and student capability. By conceptualizing and formulating Data-Model Compatibility (DMC), our work provides a quantitative and interpretable means to evaluate dataset suitability for specific student models. Empirical results confirm that DMC serves as a stronger predictor of downstream reasoning performance than conventional measures. The functional form of DMC indicates that, in reasoning distillation, higher data quality is naturally preferable; however, the level of data difficulty that best fits the model varies with the model’s capability.
On the other hand, dynamic data selection based on DMC can further improve the performance. The dynamic data selection scheme is analogous to human curriculum design: a set of core courses represents essential knowledge that the model must continually learn and retain, while elective courses are dynamically adjusted to further refine the model’s reasoning ability. Furthermore, the gains transfer to out-of-distribution tasks, suggesting that DMC captures a generalizable signal of learnability rather than a domain-specific heuristic.

\section*{Limitations}
Although our proposed DMC demonstrates promising results in data selection for reasoning distillation, several limitations remain to be addressed in future work:

\textbf{Computational efficiency.} The dynamic data selection based on DMC requires an additional re-evaluation stage after each training epoch, resulting in an approximately 53\% increase in training time, which may limit scalability when applied to massive datasets or large student models. However, this overhead is largely optimizable: since low-quality data are rarely selected (Figure \ref{fig:distribution selected data}, Section \ref{sec: discussion}), the re-evaluation can be restricted to the top 20\% highest-quality candidates, reducing the additional training-time overhead from $\sim$53\% to $\sim$10.6\% (see Appendix \ref{app: cost}). Given that distillation targets inference efficiency, this marginal one-time cost is an acceptable trade-off for a stronger student model.

\textbf{Scope of focus.} Our current work primarily concentrates on selecting compatible data instances from existing reasoning datasets rather than generating new ones. In future work, we plan to extend DMC as a guiding metric to assist teacher models in actively generating reasoning data that better aligns with the student model’s capabilities, thereby unifying data generation and selection under a single adaptive framework.

\textbf{Domain generalization.} Our primary experiments focus on the reasoning distillation domain. Beyond the in-domain benchmarks, we additionally verify cross-domain transfer on out-of-distribution tasks (Section \ref{sec: ood}); extending and validating DMC for other synthetic-data-centric settings, such as instruction tuning, dialogue generation, or data-centric LLM optimization, remains a promising direction.

\textbf{Theoretical grounding.} Rather than assuming a fixed heuristic (e.g., ``harder is better''), DMC is \emph{discovered} from data via symbolic regression, and a Pareto-frontier selection mitigates the risk of arbitrary or overfitted forms. Nonetheless, the resulting formula is empirically derived rather than analytically proven. We view its interpretable structure and consistent behavior across models and tasks as empirical evidence and a foundation for future theoretical investigations into capability-conditioned data evaluation.




\bibliography{custom}

\appendix

\section{Prompt Templates}
\label{app: prompt}

Prompt templates used for finetuning are given in Table \ref{tab: prompt}.

\begin{table}[tbh]
\centering
\small
\begin{tabular}{p{0.45\textwidth}}
\toprule
\textbf{Prompt Template} \\
\midrule
Answer the following question: \#\#\# Question: \{question\} \#\#\# Answer: \{reasoning\}.\\

\bottomrule
\end{tabular}
\caption{Prompt templates used in finetuning. During training, the loss was computed only for the reasoning and the answer.}
\label{tab: prompt}
\end{table}

\section{Student Capability}
The relative student capability of student models on different tasks are given in Table \ref{tab:capability-model-task}.
\begin{table*}[htb]
\centering
\adjustbox{max width=0.8\textwidth}{
\begin{tabular}{
                >{\raggedright\arraybackslash}m{1.5cm}|
                >{\centering\arraybackslash}m{1.2cm}
                >{\centering\arraybackslash}m{1.2cm}
                >{\centering\arraybackslash}m{1.2cm}
                >{\centering\arraybackslash}m{1.2cm}
                >{\centering\arraybackslash}m{1.2cm}
                >{\centering\arraybackslash}m{1.2cm}
                >{\centering\arraybackslash}m{1.2cm}
                >{\centering\arraybackslash}m{1.2cm}
                >{\centering\arraybackslash}m{1.2cm}}
\toprule
\multirow{2}{*}{\makecell[c]{\textbf{Task}} }& \multicolumn{8}{c}{\textbf{Student Models} $S$}\\
&G-2b&M-7b&Q-1.5b&Q-3b&Q-7b&L-1b&L-3b&L-8b \\
\midrule
SQA & 0.075 & 0.157 & 0.622 & 0.651 & 0.603 & 0.675 & 0.695 & 0.683 \\
CSQA & 0.384 & 0.531 & 0.918 & 1.000 & 0.926 & 0.919 & 0.967 & 0.975 \\
ARC & 0.390 & 0.464 & 0.694 & 0.729 & 0.690 & 0.722 & 0.756 & 0.755 \\
MATH & 0.000 & 0.118 & 0.377 & 0.411 & 0.380 & 0.438 & 0.452 & 0.440 \\
GSM8K & 0.079 & 0.190 & 0.540 & 0.595 & 0.544 & 0.676 & 0.708 & 0.668 \\
\midrule
\textbf{Average} & 0.186 & 0.292 & 0.630 & 0.677 & 0.629 & 0.686 & 0.716 & 0.704 \\
\bottomrule
\end{tabular}
}
\caption{Capability of student models across different tasks.}
\label{tab:capability-model-task}
\end{table*}

\section{Details for Data-Model Compatibility}
\subsection{Psuedocode for DMC Optimization}
To obtain the optimal configuration for $\text{DMC}$, we followed the steps below, as shown in Algorithm \ref{alg:dmc_optimization}: 
(1) Enumerate all possible combinations of evaluation metrics for $Q$, $D$, and $C$; (2) For each combination, employ symbolic regression to derive the functional forms of $M_S^L(\cdot)$, $M_S^H(\cdot)$ and $f(C_S)$; (3) Manually filter the resulting equations to retain those with stronger interpretability and better performance; (4) Apply grid search to finetune the parameters within the candidates for $M_S^L(\cdot)$ and $M_S^H(\cdot)$ and identify the optimal configuration among all combinations.
\label{app: psuedocode}
\begin{algorithm*}[tbh]
\caption{DMC Optimization and Evaluation Strategy}
\label{alg:dmc_optimization}
\begin{algorithmic}[1]
\REQUIRE Candidate metric sets $\mathbb{Q}, \mathbb{D}, \mathbb{C}$; Student model $S$; Datasets $[\mathcal{D}_i]|_i$; Ground-truth performance $[P_S(\mathcal{D}_i)]|_i$.
\ENSURE Optimal DMC configuration $(Q^*, D^*, C^*)$ and functional forms $M_S^L, M_S^H, f$.

\STATE \COMMENT{// Step 1: Enumeration of Metric Combinations}
\STATE $\mathcal{H} \leftarrow \emptyset$ \COMMENT{Candidate pool}
\FORALL{$Q_j \in \mathbb{Q}, D_k \in \mathbb{D}, C_l \in \mathbb{C}$}
    \STATE $C_S \leftarrow \text{PlacementTest}(S, C_l)$
    \STATE $Q(\mathcal{D}_i) \leftarrow \text{EvaluateQuality}(\mathcal{D}_i, Q_j)$
    \STATE $D_S(\mathcal{D}_i) \leftarrow \text{EvaluateDifficulty}(\mathcal{D}_i, D_k, S)$
    
    \STATE \COMMENT{// Step 2: Symbolic Regression for Functional Forms}
    \STATE Objective: $\max \text{Corr}([\text{DMC}_S (\mathcal{D}_i)]|_i, [P_S (\mathcal{D}_i)]|_i$
    \STATE $\{M_S^L, M_S^H, f\} \leftarrow \text{SymbolicRegression}(\text{Inputs}=\{Q, D_S, C_S\})$
    \STATE Store $(Q_j, D_k, C_l, M_S^L, M_S^H, f, \text{Score})$ in $\mathcal{H}$
\ENDFOR

\STATE \COMMENT{// Step 3: Interpretability Filtering}
\STATE $\mathcal{H}_{filtered} \leftarrow \{h \in \mathcal{H} \mid h \text{ is human-interpretable and adheres to Occam's Razor}\}$

\STATE \COMMENT{// Step 4: Fine-tuning via Grid Search}
\STATE $\text{BestCorr} \leftarrow -1$
\FORALL{candidate $h \in \mathcal{H}_{filtered}$}
    \STATE $\Theta^* \leftarrow \arg\max_{\Theta} \text{Corr}(\text{DMC}_S(\mathcal{D}_i; \Theta, h), P_{S, \mathcal{D}_i})$ \COMMENT{Optimize constants}
    \IF{$\text{Corr}(h; \Theta^*) > \text{BestCorr}$}
        \STATE $\text{BestCorr} \leftarrow \text{Corr}(h; \Theta^*)$
        \STATE $(Q^*, D^*, C^*, M_S^L, M_S^H, f) \leftarrow (h, \Theta^*)$
    \ENDIF
\ENDFOR

\RETURN Best DMC configuration
\end{algorithmic}
\end{algorithm*}

\subsection{Visualization of Optimal DMC Configuration}
\label{app: visualization DMC config}
\begin{figure}
    \centering
    \includegraphics[width=1\linewidth]{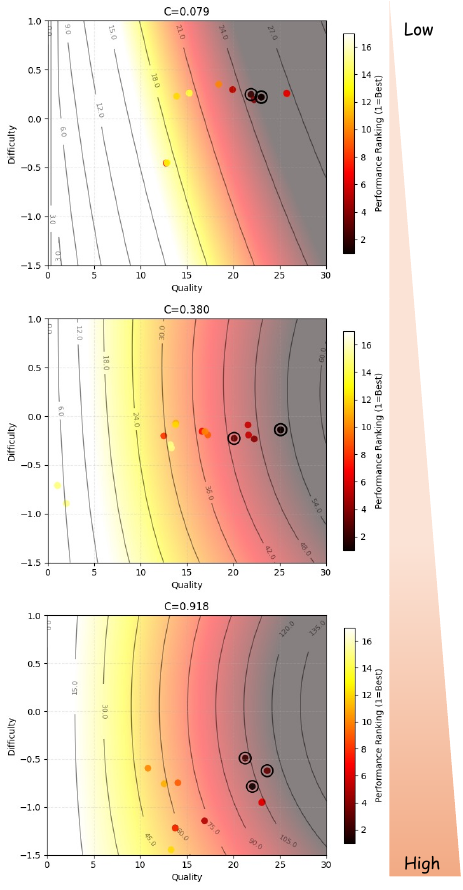}
    \caption{Model’s performance ranking across subsets $\mathcal{D}_i$ and the distribution of DMC values with student capability varying. From top to bottom, the plots represent the performance of Gemma‑2B on GSM8K, Qwen2.5‑7B on MATH, and Llama3.2‑1B on CSQA, respectively. $C$ adopts the relative student capability, taking values within the range $[0, 1]$.}
    \label{fig: dmc-capability-trend}
\end{figure}

Representative examples illustrating how the model’s performance ranking across subsets $\mathcal{D}_i$ and the distribution of DMC values vary as student capability increases are presented in Figure \ref{fig: dmc-capability-trend}.
Each point in the figure corresponds to a subset $\mathcal{D}_i$, where darker point colors indicate better distillation $P_S(\mathcal{D}_i)$ on it. The background color depicts the distribution of $\text{DMC}_S$, with darker shades signifying higher $\text{DMC}_S$ levels.

\begin{figure}[htb]
    \centering
    \includegraphics[width=1\linewidth]{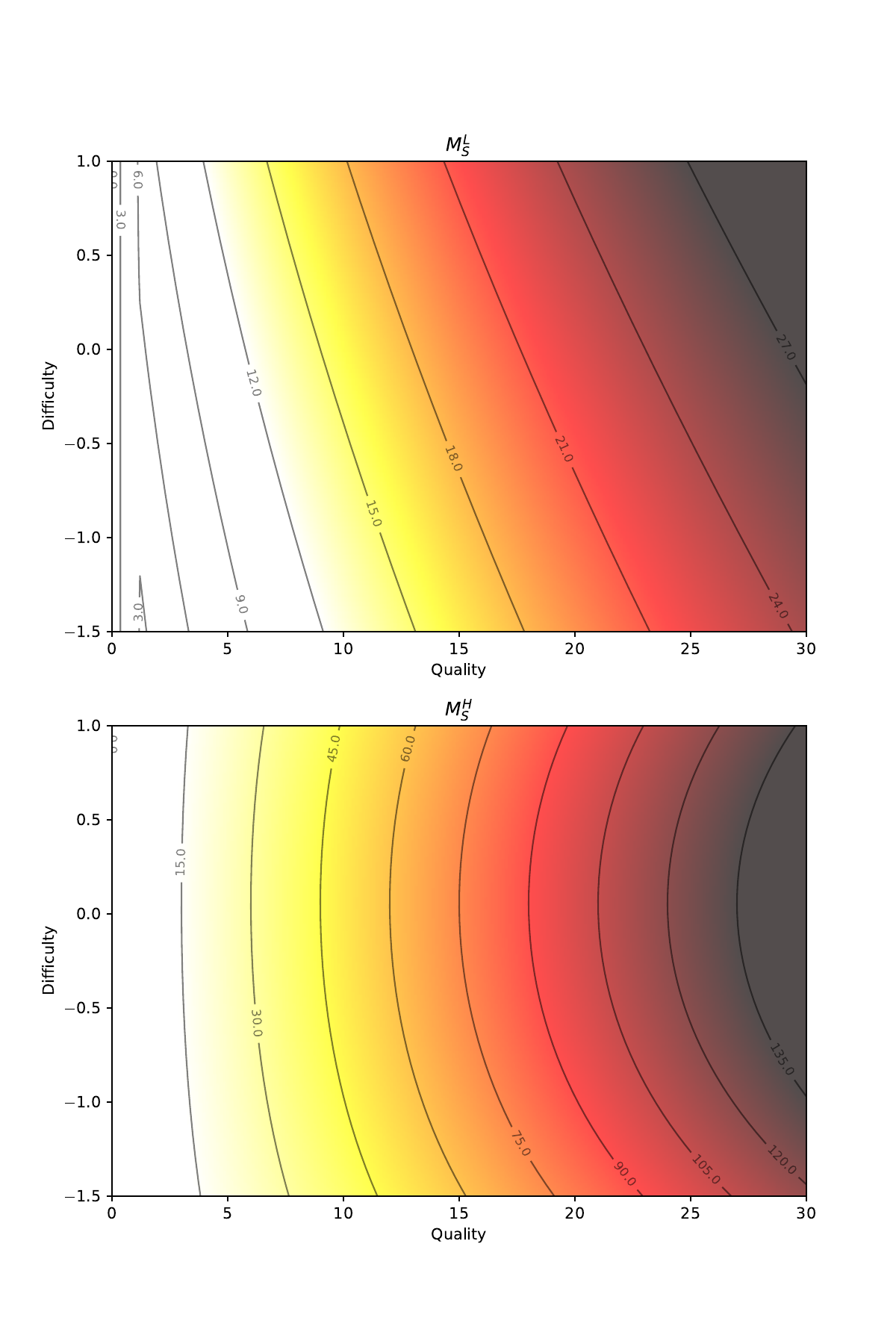}
    \caption{Visualization of $M_S^L$ and $M_S^H$.}
    \label{fig:vis msl msh}
\end{figure}

\begin{figure}
    \centering
    \includegraphics[width=1\linewidth]{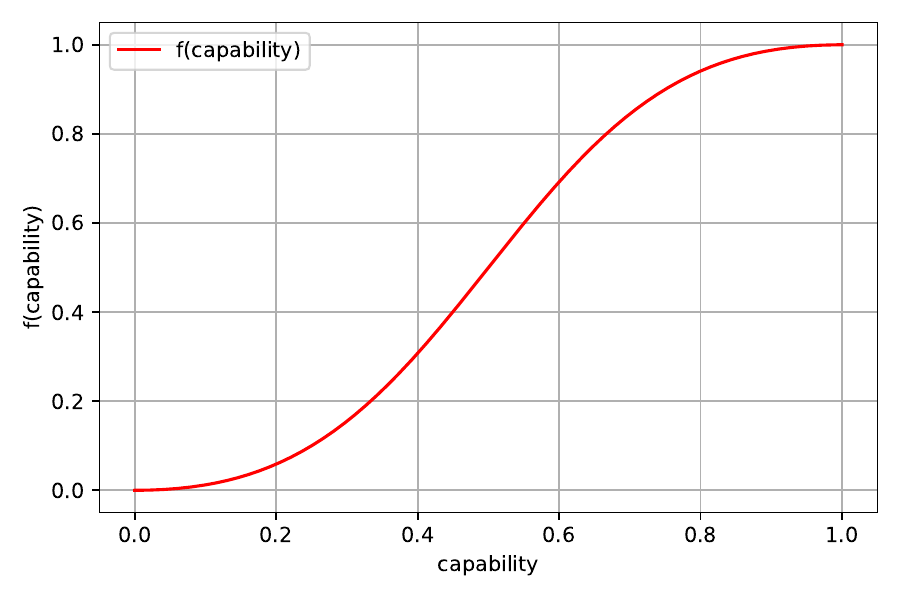}
    \caption{Visualization of $f(C_S)$}
    \label{fig:f(C_S)}
\end{figure}

The visualization of $M_S^L(\cdot)$, $M_S^H(\cdot)$ and $f(C_S)$ are given in Figure \ref{fig:vis msl msh} and \ref{fig:f(C_S)}. 

\subsection{Full Table of Correlation}
\label{sec: corr}
Table \ref{tab:corr-full} is the full table demonstrating the correlation between reasoning distillation performance and evaluation metrics across different student models and tasks.
\begin{table*}
\centering
\adjustbox{max width=0.9\textwidth}{
\begin{tabular}{>{\raggedright\arraybackslash}m{1.8cm}
                >{\raggedright\arraybackslash}m{0.9cm}|
                >{\centering\arraybackslash}m{1.2cm}
                >{\centering\arraybackslash}m{1.2cm}
                >{\centering\arraybackslash}m{1.2cm}
                >{\centering\arraybackslash}m{1.2cm}
                >{\centering\arraybackslash}m{1.2cm}
                >{\centering\arraybackslash}m{1.2cm}
                >{\centering\arraybackslash}m{1.2cm}
                >{\centering\arraybackslash}m{1.2cm}|
                >{\centering\arraybackslash}m{1.2cm}}
\toprule
\multicolumn{2}{c|}{\multirow{2}{*}{\makecell[c]{\textbf{Metric} $M$} }}& \multicolumn{8}{c|}{\textbf{Student Models} $S$}& \multirow{2}{*}{\makecell[c]{\textbf{Avg.}}} \\
&&G-2b&M-7b&Q-1.5b&Q-3b&Q-7b&L-1b&L-3b&L-8b& \\
\midrule
\multicolumn{11}{c}{\textbf{SQA}} \\
\midrule
\multirow{2}{*}{\makecell[c]{\textbf{Quality}}}
&\textbf{SRL}&0.357 & 0.119 & 0.741 & 0.370 & 0.444 & 0.427 & 0.509 & 0.484 & 0.431 \\
&\textbf{SRG}&0.268 & 0.083 & 0.635 & 0.536 & 0.257 & 0.390 & 0.330 & 0.429 & 0.366 \\

\midrule
\multirow{3}{*}{\makecell[c]{\textbf{Difficulty}}}
&\textbf{PPL}&-0.404 & 0.186 & -0.542 & -0.355 & -0.307 & -0.678 & -0.671 & -0.309 & -0.385 \\
&\textbf{CPPL}&0.427 & -0.127 & 0.729 & 0.651 & 0.107 & 0.450 & 0.630 & 0.389 & 0.407 \\
&\textbf{IFD}&0.450 & -0.113 & 0.804 & 0.575 & 0.141 & 0.382 & 0.615 & 0.184 & 0.380 \\

\midrule
\multirow{2}{*}{\makecell[c]{\textbf{Multi-Metric}}}
&\textbf{CAR}&0.188 & 0.018 & 0.802 & 0.409 & 0.342 & 0.410 & 0.515 & 0.448 & 0.392 \\
&\textbf{DMC}&0.425&0.204&0.773& 0.433& 0.402&0.470&0.590&0.506 & 0.475 \\

\midrule
\multicolumn{11}{c}{\textbf{CSQA}} \\
\midrule
\multirow{2}{*}{\makecell[c]{\textbf{Quality}}}
&\textbf{SRL}&0.385 & 0.673 & 0.860 & 0.694 & 0.840 & 0.747 & 0.877 & 0.064 & 0.642 \\
&\textbf{SRG}&0.382 & 0.686 & 0.627 & 0.701 & 0.730 & 0.690 & 0.532 & 0.017 & 0.546 \\

\midrule
\multirow{3}{*}{\makecell[c]{\textbf{Difficulty}}}
&\textbf{PPL}&-0.233 & -0.396 & -0.583 & -0.110 & -0.387 & -0.261 & -0.629 & -0.262 & -0.358 \\
&\textbf{CPPL}&0.176 & 0.392 & 0.733 & 0.195 & 0.506 & 0.362 & 0.655 & -0.027 & 0.374 \\
&\textbf{IFD}&0.039 & 0.349 & 0.772 & 0.417 & 0.631 & 0.070 & 0.429 & -0.632 & 0.259 \\
\midrule
\multirow{2}{*}{\makecell[c]{\textbf{Multi-Metric}}}
&\textbf{CAR}&0.145 & 0.446 & 0.868 & 0.430 & 0.701 & 0.565 & 0.876 & -0.010 & 0.503 \\
&\textbf{DMC}&0.403&0.696&0.895&0.618&0.843&0.778&0.923&0.204&0.670 \\
\midrule
\multicolumn{11}{c}{\textbf{ARC}} \\
\midrule
\multirow{2}{*}{\makecell[c]{\textbf{Quality}}}
&\textbf{SRL}&0.356 & 0.604 & 0.696 & 0.939 & 0.810 & 0.414 & 0.632 & 0.123 & 0.572 \\
&\textbf{SRG}&0.036 & 0.446 & 0.358 & 0.837 & 0.684 & 0.414 & 0.329 & 0.006 & 0.389 \\

\midrule
\multirow{3}{*}{\makecell[c]{\textbf{Difficulty}}}
&\textbf{PPL}&-0.828 & -0.756 & -0.542 & -0.399 & -0.498 & -0.123 & -0.674 & -0.562 & -0.548 \\
&\textbf{CPPL}&0.836 & 0.800 & 0.453 & 0.295 & 0.338 & -0.704 & -0.487 & -0.205 & 0.166 \\
&\textbf{IFD}&0.797 & 0.714 & 0.233 & 0.269 & 0.275 & -0.399 & -0.755 & -0.364 & 0.096 \\

\midrule
\multirow{2}{*}{\makecell[c]{\textbf{Multi-Metric}}}
&\textbf{CAR}&0.761 & 0.788 & 0.566 & 0.718 & 0.727 & 0.188 & 0.517 & 0.218 & 0.560 \\
&\textbf{DMC}&0.598&0.752&0.780&0.901&0.828&0.476&0.718&0.234&0.661 \\
\midrule
\multicolumn{11}{c}{\textbf{MATH}} \\
\midrule
\multirow{2}{*}{\makecell[c]{\textbf{Quality}}}
&\textbf{SRL}&0.363 & 0.121 & 0.602 & 0.527 & 0.900 & 0.831 & 0.600 & 0.556 & 0.562 \\
&\textbf{SRG}&0.321 & 0.101 & 0.608 & 0.507 & 0.890 & 0.833 & 0.627 & 0.607 & 0.562 \\

\midrule
\multirow{3}{*}{\makecell[c]{\textbf{Difficulty}}}
&\textbf{PPL}&-0.773 & -0.682 & -0.878 & -0.931 & -0.656 & -0.696 & -0.934 & -0.823 & -0.797 \\
&\textbf{CPPL}&0.759 & 0.502 & 0.434 & 0.456 & 0.514 & 0.844 & 0.776 & 0.705 & 0.624 \\
&\textbf{IFD}&0.770 & 0.661 & -0.793 & -0.794 & -0.648 & 0.700 & 0.440 & 0.421 & 0.095 \\

\midrule
\multirow{2}{*}{\makecell[c]{\textbf{Multi-Metric}}}
&\textbf{CAR}&-0.281 & 0.161 & 0.540 & 0.588 & 0.915 & 0.901 & 0.668 & 0.680 & 0.521 \\
&\textbf{DMC}&0.469&0.306&0.659&0.594&0.900&0.861&0.711&0.684&0.648\\
\midrule
\multicolumn{11}{c}{\textbf{GSM8K}} \\
\midrule
\multirow{2}{*}{\makecell[c]{\textbf{Quality}}}
&\textbf{SRL}&0.800 & 0.685 & 0.351 & 0.748 & 0.822 & -0.016 & 0.690 & 0.377 & 0.557 \\
&\textbf{SRG}&0.155 & 0.230 & 0.464 & 0.480 & 0.281 & 0.364 & 0.551 & 0.583 & 0.388 \\

\midrule
\multirow{3}{*}{\makecell[c]{\textbf{Difficulty}}}
&\textbf{PPL}&-0.329 & -0.667 & -0.839 & -0.763 & -0.378 & -0.588 & -0.702 & -0.831 & -0.637 \\
&\textbf{CPPL}&0.299 & 0.599 & -0.363 & -0.585 & -0.791 & 0.424 & 0.620 & 0.839 & 0.130 \\
&\textbf{IFD}&0.321 & 0.609 & -0.528 & -0.716 & -0.521 & -0.501 & -0.797 & -0.659 & -0.349 \\

\midrule
\multirow{2}{*}{\makecell[c]{\textbf{Multi-Metric}}}
&\textbf{CAR}&0.515 & 0.626 & -0.221 & 0.418 & 0.227 & -0.010 & 0.680 & 0.382 & 0.327 \\
&\textbf{DMC}&0.800&0.735& 0.401&0.798&0.822&0.143& 0.735&0.425&0.607 \\

\bottomrule
\end{tabular}
}
\caption{The correlation between reasoning distillation performance and evaluation metrics across different student models and tasks.}
\label{tab:corr-full}
\end{table*}

\subsection{Relation between $P$, $Q$ and $D$}
\label{sec: relation pqd}
\begin{figure*}
    \centering
    \includegraphics[width=1\linewidth]{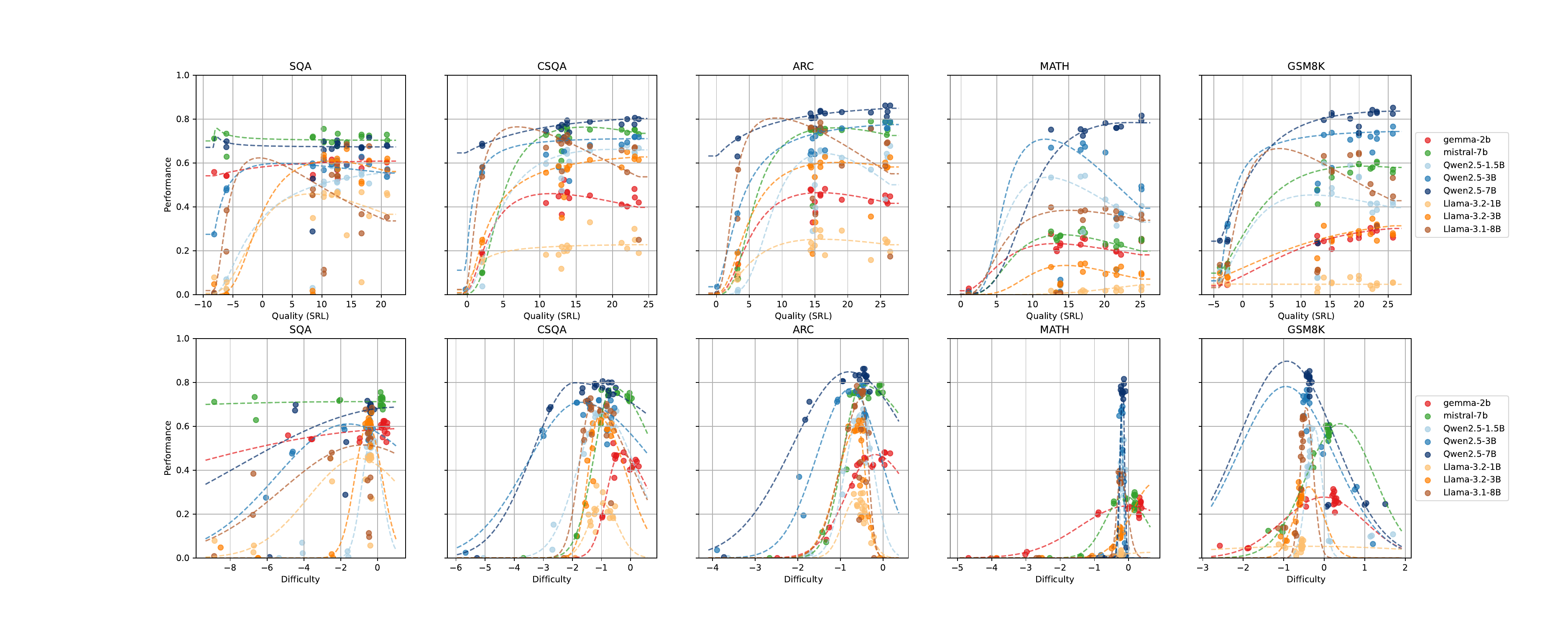}
    \caption{Scatter plots of $P$ versus $Q$ and $P$ versus $D$, along with the corresponding regression lines.}
    \label{fig:relation pqd}
\end{figure*}

Figure \ref{fig:relation pqd} presents the relation between $P$ and $Q$  as well as between $P$ and $D$, and displays the regression curves.

\section{Reasoning Distillation Performance}
\label{app: performance}
Table \ref{tab:performance-g-2b}, \ref{tab:performance-q-1.5b}, \ref{tab:performance-q-7b} shows the performance of multiple data selection methods on the tasks, with student model $S$ selected as Gemma-2B, Mistral-7B, Qwen2.5-1.5B and Qwen2.5-7B. Among them, Gemma‑2B represents a student model with relatively low initial capability and limited potential; Mistral‑7B denotes a model with low initial reasoning ability but high potential owing to its larger parameter count; and the Qwen family is primarily used to compare the performance of models within the same family across different parameter scales.

\begin{table}
\centering
\adjustbox{max width=0.5\textwidth}{
\begin{tabular}{
>{\raggedright\arraybackslash}m{1.8cm}|
>{\raggedright\arraybackslash}m{3.0cm}|
>{\centering\arraybackslash}m{1.2cm}
>{\centering\arraybackslash}m{1.2cm}
>{\centering\arraybackslash}m{1.2cm}
>{\centering\arraybackslash}m{1.2cm}
>{\centering\arraybackslash}m{1.2cm}}
\toprule
\multirow{2}{*}{\makecell[c]{\makecell[l]{\textbf{Training}\\\textbf{Dataset}}}} & \multirow{2}{*}{\makecell[c]{\makecell[l]{\textbf{Selection Metric}}\\\textbf{(Top $k\%$)}}} & \multicolumn{5}{c}{\textbf{Tasks}}\\
& & SQA & CSQA & ARC & MATH & GSM8K \\
\midrule
\textbf{None} & \makecell[c]{-} & 48.2 & 38.5 & 40.1 & 8.4 & 12.6 \\
\midrule
\multirow{9}{*}{\makecell[l]{\textbf{Static}}}
& Full (100\%)  & 54.6 & 45.2 & 44.8 & 18.2 & 23.4 \\
& Random (12.5\%) & 51.4 & 43.8 & 42.6 & 14.5 & 19.8 \\
& Quality (12.5\%) & 56.2 & 46.8 & 45.4 & \underline{25.4} & 24.6 \\
& Difficulty (12.5\%) & 49.8 & 44.5 & 41.2 & 16.8 & 18.2 \\
& CAR (12.5\%)  & 53.5 & 47.2 & 46.0 & 22.0 & 25.2 \\
& DMC (12.5\%)  & \underline{58.4} & \textbf{49.6} & \textbf{47.5} & \textbf{26.8} & \underline{28.4} \\
& DMC (10\%)   & \textbf{59.2} & \underline{48.4} & \underline{46.8} & \underline{25.4} & \textbf{29.1} \\
& DMC (7.5\%)  & 57.8 & 47.9 & 46.2 & 23.6 & 27.5 \\
\midrule
\multirow{6}{*}{\textbf{Dynamic}}
& ADCL  & 55.4 & 48.2 & 43.5 & 20.6 & 22.8 \\
& Difficulty (12.5\%) & 50.2 & 46.5 & 42.8 & 19.4 & 21.2 \\
& CAR (12.5\%) & 56.8 & 50.4 & 47.2 & 26.5 & 29.5 \\
& DMC (12.5\%) & \underline{61.6} & \textbf{55.2} & 48.2 & \textbf{29.6} & 30.8 \\
& DMC (10\%)  & \textbf{62.5} & \underline{52.3} & \textbf{49.4} & \underline{28.6} & \textbf{32.4} \\
& DMC (7.5\%) & 60.3 & 51.0 & \underline{49.2} & 27.8 & \underline{31.8} \\
\bottomrule
\end{tabular}
}
\caption{Performance of Gemma-2B. For the static and dynamic settings, the dataset selection metric achieving the best and second-best performance are indicated in \textbf{bold} and \underline{underline}, respectively.}
\label{tab:performance-g-2b}
\end{table}

\begin{table}
\centering
\adjustbox{max width=0.5\textwidth}{
\begin{tabular}{
>{\raggedright\arraybackslash}m{1.8cm}|
>{\raggedright\arraybackslash}m{3.0cm}|
>{\centering\arraybackslash}m{1.2cm}
>{\centering\arraybackslash}m{1.2cm}
>{\centering\arraybackslash}m{1.2cm}
>{\centering\arraybackslash}m{1.2cm}
>{\centering\arraybackslash}m{1.2cm}}
\toprule
\multirow{2}{*}{\makecell[c]{\makecell[l]{\textbf{Training}\\\textbf{Dataset}}}} & \multirow{2}{*}{\makecell[c]{\makecell[l]{\textbf{Selection Metric}}\\\textbf{(Top $k\%$)}}} & \multicolumn{5}{c}{\textbf{Tasks}}\\
& & SQA & CSQA & ARC & MATH & GSM8K \\
\midrule
\textbf{None} & \makecell[c]{-} & 65.2 & 64.8 & 69.4 & 10.5 & 35.2 \\
\midrule
\multirow{9}{*}{\makecell[l]{\textbf{Static}}}
& Full (100\%)  & 69.4 & 71.6 & 75.0 & 23.0 & 47.8 \\
& Random (12.5\%)  & 67.8 & 70.2 & 73.5 & 19.8 & 45.4 \\
& Quality (12.5\%)  & \textbf{72.6} & 73.5 & 76.8 & 26.5 & 54.2 \\
& Difficulty (12.5\%)  & 66.2 & 68.4 & 70.2 & 14.2 & 41.5 \\
& CAR (12.5\%)   & 70.5 & 73.2 & 76.2 & 25.8 & 53.8 \\
& DMC (12.5\%)   & \underline{72.5} & \textbf{75.4} & \textbf{78.2} & \textbf{29.4} & \underline{58.6} \\
& DMC (10\%)  & 71.8 & \underline{74.8} & \underline{77.5} & \underline{28.2} & \textbf{59.2} \\
& DMC (7.5\%)  & 70.2 & 73.6 & 76.8 & 26.0 & 56.4 \\
\midrule
\multirow{6}{*}{\textbf{Dynamic}}
& ADCL   & 70.8 & 72.5 & 74.2 & 22.4 & 51.0 \\
& Difficulty (12.5\%) & 65.4 & 70.6 & 72.8 & 20.2 & 48.6 \\
& CAR (12.5\%)  & 71.6 & 75.2 & 78.4 & 28.5 & 59.5 \\
& DMC (12.5\%)  & \underline{73.9} & \underline{76.6} & \textbf{80.2} & \textbf{31.2} & \underline{62.4} \\
& DMC (10\%)  & \textbf{75.3} & \textbf{77.2} & \underline{79.0} & \underline{30.0} & \textbf{62.6} \\
& DMC (7.5\%)  & 73.7 & 76.2 & \underline{79.0} & 30.6 & 60.6 \\
\bottomrule
\end{tabular}
}
\caption{Performance of Mistral-7B. For the static and dynamic settings, the dataset selection metric achieving the best and second-best performance are indicated in \textbf{bold} and \underline{underline}, respectively.}
\label{tab:performance-m-7b}
\end{table}

\begin{table}
\centering
\adjustbox{max width=0.5\textwidth}{
\begin{tabular}{
>{\raggedright\arraybackslash}m{1.8cm}|
>{\raggedright\arraybackslash}m{3.0cm}|
>{\centering\arraybackslash}m{1.2cm}
>{\centering\arraybackslash}m{1.2cm}
>{\centering\arraybackslash}m{1.2cm}
>{\centering\arraybackslash}m{1.2cm}
>{\centering\arraybackslash}m{1.2cm}}
\toprule
\multirow{2}{*}{\makecell[c]{\makecell[l]{\textbf{Training}\\\textbf{Dataset}}}} & \multirow{2}{*}{\makecell[c]{\makecell[l]{\textbf{Selection Metric}}\\\textbf{(Top $k\%$)}}} & \multicolumn{5}{c}{\textbf{Tasks}}\\
& & SQA & CSQA & ARC & MATH & GSM8K \\
\midrule
\textbf{None} & \makecell[c]{-} & 45.2 & 38.4 & 52.6 & 9.4 & 14.2 \\
\midrule
\multirow{9}{*}{\makecell[l]{\textbf{Static}}}
& Full (100\%) & 51.5 & 59.6 & 61.6 & 31.4 & 41.6 \\
& Random (12.5\%) & 50.1 & 58.2 & 59.4 & 28.6 & 38.5 \\
& Quality (12.5\%) & \textbf{54.8} & 62.5 & 63.8 & 44.2 & 51.4 \\
& Difficulty (12.5\%) & 48.6 & 57.0 & 55.2 & 18.4 & 32.0 \\
& CAR (12.5\%)  & 52.4 & 61.8 & 63.2 & 41.0 & 50.8 \\
& DMC (12.5\%)  & \underline{54.2} & \textbf{65.4} & \textbf{65.8} & \textbf{51.2} & \underline{55.6} \\
& DMC (10\%) & 53.8 & \underline{64.8} & \underline{65.2} & \underline{49.8} & \textbf{56.4} \\
& DMC (7.5\%)  & 52.6 & 63.2 & 64.0 & 46.5 & 53.2 \\
\midrule
\multirow{6}{*}{\textbf{Dynamic}}
& ADCL & 52.8 & 60.5 & 58.6 & 35.2 & 42.4 \\
& Difficulty (12.5\%) & 47.4 & 58.4 & 56.8 & 30.2 & 35.6 \\
& CAR (12.5\%) & 53.2 & 64.2 & 64.6 & 48.4 & 54.8 \\
& DMC (12.5\%) & \underline{55.9} & \textbf{69.2} & 67.2 & \underline{53.6} & \textbf{59.8} \\
& DMC (10\%) & \textbf{56.2} & \underline{68.0} & \textbf{67.8} & \textbf{54.2} & \underline{59.0} \\
& DMC (7.5\%) & 54.3 & 67.0 & \underline{65.4} & 53.8 & 57.2 \\
\bottomrule
\end{tabular}
}
\caption{Performance of Qwen2.5-1.5B. For the static and dynamic settings, the dataset selection metric achieving the best and second-best performance are indicated in \textbf{bold} and \underline{underline}, respectively.}
\label{tab:performance-q-1.5b}
\end{table}

\begin{table}
\centering
\adjustbox{max width=0.5\textwidth}{
\begin{tabular}{
>{\raggedright\arraybackslash}m{1.8cm}|
>{\raggedright\arraybackslash}m{3.0cm}|
>{\centering\arraybackslash}m{1.2cm}
>{\centering\arraybackslash}m{1.2cm}
>{\centering\arraybackslash}m{1.2cm}
>{\centering\arraybackslash}m{1.2cm}
>{\centering\arraybackslash}m{1.2cm}}
\toprule
\multirow{2}{*}{\makecell[c]{\makecell[l]{\textbf{Training}\\\textbf{Dataset}}}} & \multirow{2}{*}{\makecell[c]{\makecell[l]{\textbf{Selection Metric}}\\\textbf{(Top $k\%$)}}} & \multicolumn{5}{c}{\textbf{Tasks}}\\
& & SQA & CSQA & ARC & MATH & GSM8K \\
\midrule
\textbf{None} & \makecell[c]{-} & 62.4 & 68.2 & 74.5 & 35.8 & 48.6 \\
\midrule
\multirow{9}{*}{\makecell[l]{\textbf{Static}}}
& Full (100\%) & 66.8 & 75.4 & 80.6 & 76.0 & 73.6 \\
& Random (12.5\%) & 64.2 & 73.8 & 78.4 & 70.2 & 68.5 \\
& Quality (12.5\%) & \underline{70.8} & \textbf{78.8} & 82.4 & 77.8 & 75.4 \\
& Difficulty (12.5\%) & 60.5 & 72.1 & 72.8 & 45.6 & 55.2 \\
& CAR (12.5\%)  & 65.8 & 76.8 & 81.6 & 76.4 & 74.8 \\
& DMC (12.5\%)  & \textbf{71.2} & \underline{78.6} & \textbf{84.2} & \textbf{80.4} & \underline{79.2} \\
& DMC (10\%) & \underline{70.8} & 78.2 & \underline{83.5} & \underline{79.6} & \textbf{80.6} \\
& DMC (7.5\%)  & 68.5 & 77.4 & 82.0 & 75.2 & 77.4 \\
\midrule
\multirow{6}{*}{\textbf{Dynamic}}
& ADCL & 68.2 & 76.4 & 75.2 & 62.8 & 70.4 \\
& Difficulty (12.5\%) & 62.4 & 74.6 & 73.5 & 58.2 & 62.8 \\
& CAR (12.5\%) & 69.4 & 78.5 & 82.8 & 79.5 & 81.2 \\
& DMC (12.5\%) & \textbf{73.6} & \textbf{81.2} & \underline{87.6} & \underline{82.8} & \underline{84.2} \\
& DMC (10\%) & \underline{73.2} & \underline{80.6} & \textbf{88.2} & \textbf{84.0} & \textbf{86.2} \\
& DMC (7.5\%) & 71.6 & 79.8 & 86.8 & 81.6 & 82.8 \\
\bottomrule
\end{tabular}
}
\caption{Performance of Qwen2.5-7B. For the static and dynamic settings, the dataset selection metric achieving the best and second-best performance are indicated in \textbf{bold} and \underline{underline}, respectively.}
\label{tab:performance-q-7b}
\end{table}

\section{Training Dynamics}
\label{app: training dynamics}
We analyze the training dynamics with MATH as $\mathcal{D}_0$, using Qwen2.5-3B as a representative high-capacity model and Gemma-2B as a representative low-capacity model.

\begin{figure}
    \centering
    \includegraphics[width=1\linewidth]{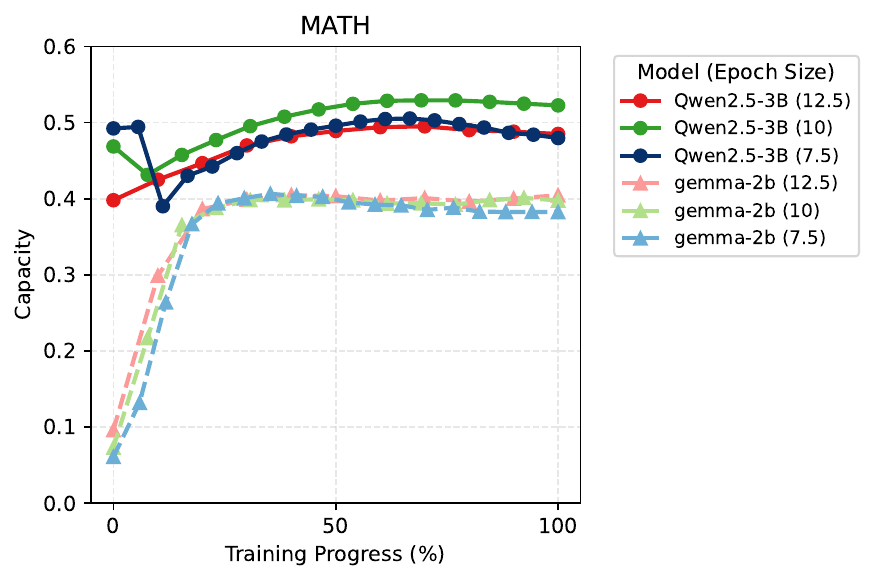}
    \caption{Student capability improvement curve under different parameter settings during reasoning distillation.}
    \label{fig: capacity shift}
\end{figure}

\paragraph{Capability Improvement}
As shown in Figure \ref{fig: capacity shift}, the student capability gradually improves as training progresses, consistent with our expectations. For the low-capability model Gemma-2B, the model initially selects relatively difficult data, yielding a rapid capability increase early in training; as its capability grows, DMC places greater emphasis on data quality and its alignment with the model, enhancing the stability of the model's reasoning capability.

\begin{figure}
    \centering
    \includegraphics[width=1\linewidth]{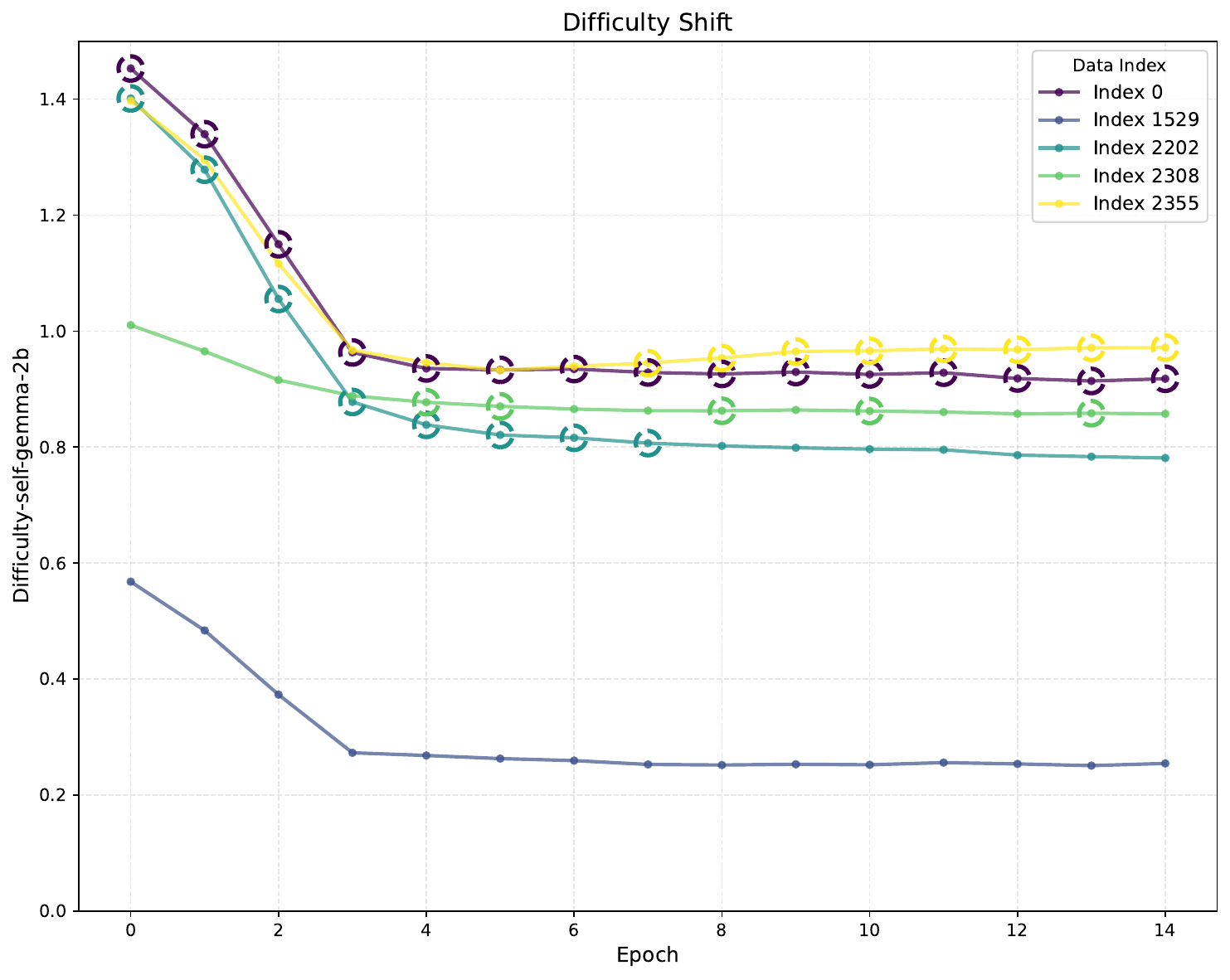}
    \caption{Difficulty shift curve of Gemma-2b on MATH.}
    \label{fig:diffictuly shift}
\end{figure}

\paragraph{Difficulty Shift}
In order to analyze changes in specific data samples, we selected five representative samples and visualized their selection status and corresponding difficulty shifts in Figure \ref{fig:diffictuly shift}. The x-axis corresponds to the training epoch progress and the y-axis to the relative difficulty. Data selected into the training set at each epoch are circled.

These samples illustrate several characteristic patterns. A core-course sample (Index 0) is consistently retained throughout training; a sample whose difficulty initially matches the student (Index 2202) is rapidly mastered and then dropped to avoid overfitting, whereas one that is initially too hard (Index 2355) is selected only once its difficulty falls into the effective DMC range, and a sample near the threshold (Index 2308) alternates in and out, regulating generalization. Notably, some excluded samples (e.g., Index 1529, 2202) keep decreasing in difficulty even without being trained on, indicating that they are not irreplaceable: the student acquires similar feature representations from other instances, which also explains why they are not selected.

\section{Computational Cost Analysis}
\label{app: cost}
Dynamic data selection re-evaluates the data pool before every training epoch, which introduces additional overhead relative to a conventional one-pass training pipeline. In our experiments, this naive re-evaluation increases the total training time by approximately 53\%.

This overhead can be substantially reduced without affecting the selected data. As shown in Figure \ref{fig:distribution selected data} (Section \ref{sec: discussion}), the dynamic selection process almost exclusively draws from the top-quality region of the pool: the highest-quality 5\% of data are nearly always selected, and data beyond the top 20\% are seldom chosen. Consequently, the expensive difficulty re-evaluation can be restricted to the top 20\% of candidates ranked by the static quality score $Q$, while the remaining low-quality data are excluded a priori. This reduces the additional training-time overhead from $\sim$53\% to $\sim$10.6\%.

We argue that this marginal one-time training cost is acceptable. The goal of distillation is to obtain an efficient student model that will serve a large number of inference calls; a $\sim$10.6\% increase in one-time training cost is a worthwhile trade-off for a consistently stronger student. Moreover, compared with the compute wasted on training over incompatible data that yields no gain, DMC-guided selection is arguably more efficient in terms of performance per unit of compute.

\end{document}